\documentclass[letterpaper, 10pt, conference]{ieeeconf}
\IEEEoverridecommandlockouts
\usepackage{graphics}
\usepackage{epsfig}
\usepackage{mathptmx}
\usepackage{times}
\usepackage{amsmath,amssymb,amsfonts}
\usepackage{graphicx}
\usepackage{booktabs}
\usepackage{multirow}
\usepackage{array}
\usepackage{xspace}
\usepackage{cite}
\usepackage{subcaption}
\newcommand{\cN}{\mathcal{N}}
\newcommand{\cA}{\mathcal{A}}
\newcommand{\cO}{\mathcal{O}}

\newcommand{\cS}{\mathcal{S}}

\newcommand{\R}{\mathbb{R}}
\newcommand{\PRISM}{\textsc{Prism}\xspace}
\newcommand{\PCTL}{\textsc{Pctl}\xspace}
\newcommand{\VQVIB}{\textsc{VQ-VIB}\xspace}

\newtheorem{proposition}{Proposition}
\newtheorem{remark}{Remark}

\newenvironment{IEEEkeywords}{\normalfont%
\begin{center}{\bf Index Terms}\end{center}%
\vspace{-3mm}\noindent\itshape}{\relax\vspace{0.67ex}}

\title{\LARGE \bf
Formal Verification of Learned Multi-Agent Communication Policies via Decision Tree Distillation
}

\author{Ahmad Farooq$^{*}$ and Kamran Iqbal% <-this % stops a space
\thanks{\scriptsize \copyright 2026 IEEE. Personal use of this material is permitted. Permission from IEEE must be obtained for all other uses, in any current or future media, including reprinting/republishing this material for advertising or promotional purposes, creating new collective works, for resale or redistribution to servers or lists, or reuse of any copyrighted component of this work in other works. This work has been accepted for publication in the \textit{2026 IEEE/RSJ International Conference on Intelligent Robots and Systems (IROS 2026)}, Pittsburgh, Pennsylvania, USA, September 27--October 1, 2026.}%
\thanks{\scriptsize A. Farooq and K. Iqbal are with the Department of Electrical and Computer Engineering, University of Arkansas at Little Rock, Little Rock, AR 72204 USA.\newline
$^{*}$Corresponding author: A. Farooq (e-mail: afarooq@ualr.edu); K. Iqbal (e-mail: kxiqbal@ualr.edu).}%
\thanks{\scriptsize ORCID: A. Farooq (0009-0002-3684-5876); K. Iqbal (0000-0001-8375-290X)}%
}

\begin{document}

\maketitle
\thispagestyle{empty}
\pagestyle{empty}

\begin{abstract}
Multi-agent reinforcement learning (MARL) enables autonomous agents to develop sophisticated coordination strategies through emergent communication, but the resulting neural network policies lack the formal safety guarantees required for deployment in safety-critical robotic applications such as drone swarms and autonomous vehicle fleets. We present the first end-to-end framework for safety verification of learned multi-agent communication policies through policy abstraction: neural policies are distilled into interpretable decision trees, which are then formally verified, with empirical validation confirming that verified safety properties transfer to the original networks. Our four-stage pipeline consists of: (1)~domain-specific feature extraction from agent observations, (2)~decision tree distillation achieving \textbf{97.9\% $\pm$ 1.2\%} fidelity to neural network policies, (3)~automated translation to \PRISM probabilistic model checker specifications with \emph{complete feature-to-state-variable correspondence}, and (4)~compositional verification of Probabilistic Computation Tree Logic (\PCTL) properties via pairwise decomposition with union-bound aggregation and empirical neighbor modeling. Evaluating Vector-Quantized Variational Information Bottleneck (\VQVIB) policies for multi-drone coordination with 5--7 agents, we verify 18 temporal logic properties across safety, liveness, and cooperation categories, achieving \textbf{88.9\% property satisfaction} with \textbf{all five safety thresholds satisfied} (0.3\% collision probability vs.\ 1\% threshold). Monte Carlo validation of the original neural policies confirms that verified safety properties transfer with $\leq$0.6 percentage-point deviation (95\% CI). Discrete \VQVIB messages provide +11.6 to +13.6 percentage point fidelity advantages over continuous communication methods, enabling 3--4$\times$ faster verification. Our framework provides empirically validated safety verification for distilled policy abstractions, which serves as a practical bridge between deep MARL and the formal safety workflows required for multi-robot deployment.
\end{abstract}


\section{INTRODUCTION}

Deploying multi-agent reinforcement learning (MARL) in safety-critical robotics applications, such as drone swarms, autonomous vehicle fleets, demands formal guarantees beyond empirical metrics. While Multi-Agent Proximal Policy Optimization (MAPPO)~\cite{yu2022surprising} and QMIX~\cite{rashid2020monotonic} achieve remarkable coordination, neural policies cannot be formally certified. The verification challenge is two-fold: neural network verification is NP-hard~\cite{katz2017reluplex}, and multi-agent systems compound this through exponential state space growth ($|S|^N$ for $N$ agents). Garg et al.~\cite{garg2024learning} identify verification of learned communication policies as an important open challenge.

We address this through \textit{policy abstraction via decision tree distillation}, obtaining symbolic representations amenable to probabilistic model checking. Our approach extends policy extraction~\cite{bastani2018verifiable,milani2022maviper} by incorporating communication semantics and coupling distillation with a complete verification pipeline. Unlike single-agent tools~\cite{gross2022cool,gross2024safety}, our framework handles multi-agent communication through pairwise compositional decomposition with empirical transfer validation.

\textbf{Key Contributions:}
\begin{enumerate}
\item \textbf{The first end-to-end abstraction-based verification framework} for learned multi-agent communication policies, comprising domain-specific feature extraction, decision tree distillation, automated \PRISM model generation with \emph{complete feature-to-state correspondence} (Sec.~\ref{sec:translation}), and compositional Probabilistic Computation Tree Logic (\PCTL) verification via pairwise decomposition with union-bound aggregation (Sec.~\ref{sec:compositional}).
\item \textbf{High-fidelity distillation} (97.9\% $\pm$ 1.2\%) enabled by domain-specific feature engineering that provides +19.7 percentage points (pp) improvement over raw observations.
\item \textbf{Full property verification}: 18 \PCTL properties across safety, liveness, and cooperation categories, with Monte Carlo validation confirming property transfer from trees to original neural policies (Sec.~\ref{sec:montecarlo}).
\item \textbf{Discrete communication advantage for finite-state verification}: \VQVIB's discrete messages provide +11.6--13.6 pp fidelity advantage over continuous methods (CommNet, TarMAC) within current finite-state probabilistic model checking (PMC) workflows, enabling 3--4$\times$ faster verification.
\end{enumerate}

\begin{figure}[t]
\centering
\includegraphics[width=\columnwidth]{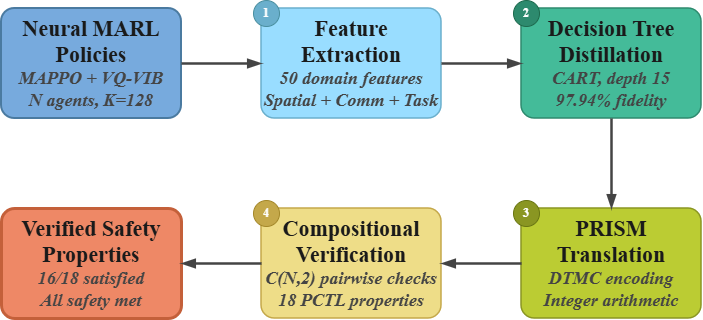}
\caption{Four-stage verification framework. Neural MARL policies are transformed through domain-specific feature extraction (50 features across spatial, communication, and task categories), decision tree distillation (97.9\% fidelity via CART), automated \PRISM translation (DTMC with complete feature-to-state correspondence), and compositional \PCTL verification ($\binom{N}{2}$ pairwise safety checks). The pipeline verifies 16/18 temporal logic properties with all safety thresholds satisfied.}
\label{fig:pipeline}
\end{figure}

\section{RELATED WORK}

\textbf{Neural Network Verification.}
Reluplex~\cite{katz2017reluplex} and DeepPoly~\cite{singh2019abstract} verify single neural networks via Satisfiability Modulo Theories (SMT) solving and abstract interpretation but do not scale to multi-agent systems with communication.
\newpage
\textbf{Policy Distillation for Verification.}
VIPER~\cite{bastani2018verifiable} extracts verifiable decision trees from single-agent RL. MAVIPER~\cite{milani2022maviper} extends this to multi-agent settings with coordinated distillation. Our work differs in two key respects: (i)~we incorporate \emph{communication features} capturing message semantics, and (ii)~we couple distillation with a complete verification pipeline. Related approaches include DTPO~\cite{liu2018toward} (direct tree training); we use Classification and Regression Trees (CART) for its simplicity and high fidelity with engineered features.

\textbf{RL Verification Pipelines.}
COOL-MC~\cite{gross2022cool} and VERINTER~\cite{gross2024safety} construct policy-induced DTMCs by querying neural networks, avoiding approximation but limited to single-agent settings. Our tree-based abstraction trades a small approximation gap for scalability to multi-agent communication. Interval DTMCs~\cite{chatterjee2008model} offer conservative bounds under transition uncertainty; integration with our pipeline is a promising direction (Sec.~\ref{sec:discussion}).

\textbf{Safe Multi-Agent Systems.}
Shielding enforces runtime safety constraints~\cite{alshiekh2018safe}, but does not verify learned behaviors. Hasanbeig et al.~\cite{hasanbeig2020cautious} incorporate temporal logic during learning. Fulton and Platzer~\cite{fulton2018safe} combine formal methods with RL for single-agent safe control.

\textbf{Compositional and Multi-Agent Verification.}
Compositional model checking~\cite{clarke1989compositional} reduces state space explosion by decomposing system verification. Kwiatkowska et al.~\cite{kwiatkowska2010assume} extend assume-guarantee reasoning to probabilistic systems with formal proof rules. PRISM-games~\cite{chen2013prism} supports verification of stochastic multi-player games but requires explicit game-theoretic models rather than learned policies. Our pairwise decomposition is inspired by these compositional techniques but uses empirically calibrated neighbor models with union-bound aggregation rather than formal assume-guarantee proof rules (Sec.~\ref{sec:compositional}). This connects to broader \emph{factored verification} approaches in multi-robot systems, where symmetry reduction and decoupled state abstractions exploit structural independence to tame state explosion~\cite{lomuscio2017mcmas}; our communication-aware pairwise decomposition extends this principle to learned policies with emergent interaction patterns.

\textbf{Multi-Agent Communication.}
CommNet~\cite{sukhbaatar2016learning} enables emergent communication through mean-pooled continuous messages. TarMAC~\cite{das2019tarmac} introduces targeted attention-based continuous communication. Recent work on bandwidth-efficient multi-agent communication~\cite{farooq2026bandwidth} combines IB with VQ. The \VQVIB architecture builds on this, pairing vector quantization~\cite{van2017neural} with variational information bottleneck~\cite{tishby2000information} for discrete messages; we verify \VQVIB policies and show that discrete communication provides clear verification advantages.

\section{PROBLEM FORMULATION}

\subsection{Multi-Agent Communication System}

We model the multi-agent system as a Decentralized Partially Observable Markov Decision Process (Dec-POMDP)~\cite{oliehoek2016concise} augmented with communication: $\langle \mathcal{I}, \cS, \{\cA^i\}, \mathcal{P}, \{R^i\}, \{\Omega^i\}, \cO, \gamma \rangle$, where $\mathcal{I} = \{1, \ldots, N\}$ is the set of agents, $\cS$ is the global state space, $\cA^i$ is agent $i$'s action space, $\mathcal{P}$ is the transition function, $R^i$ is the reward function, $\Omega^i$ is the observation space, $\cO$ is the observation function, and $\gamma$ is the discount factor.

At each timestep $t$, agent $i$ observes $o_t^i$, generates a discrete message $m_t^i \in \{1, \ldots, K\}$ through vector quantization~\cite{van2017neural}, and receives messages from neighbors $\cN_t^i = \{j : \|p_t^i - p_t^j\| \leq r_{\text{comm}}\}$. Messages are aggregated through graph attention~\cite{velivckovic2018graph}:
$\bar{z}_t^i = \sum_{j \in \cN_t^i} \alpha_{ij}^t z_t^j$,
where $\alpha_{ij}^t$ are learned attention weights and $z_t^j$ is the quantized embedding for message $m_t^j$. Agent $i$'s policy $\pi^i(a_t^i | o_t^i, \bar{z}_t^i)$ conditions on both observation and aggregated communication.

\subsection{Verification Objective}

Given trained neural policies $\{\pi^i_\theta\}_{i=1}^N$, we aim to verify whether the multi-agent system satisfies temporal logic properties $\Phi = \{\phi_1, \ldots, \phi_M\}$ expressed in \PCTL. Since direct verification of neural policies is intractable, we seek interpretable abstractions $\{\hat{\pi}^i\}$ such that: (1)~$\hat{\pi}^i$ faithfully approximates $\pi^i_\theta$ with high fidelity, and (2)~$\hat{\pi}^i$ admits efficient formal verification through established model checking tools.

\section{VERIFICATION FRAMEWORK}

Our framework consists of four stages (Fig.~\ref{fig:pipeline}).

\subsection{Stage 1: Domain-Specific Feature Extraction}

Raw agent observations (14-dimensional vectors of position, velocity, and local sensing) are insufficient for high-fidelity distillation, achieving only 78.2\% accuracy. We design 50 candidate features organized into three groups:

\textbf{Spatial Features (15):} Agent position (normalized), velocity, distances to nearest targets and neighbors, boundary distances, quadrant indicators, angular features. All computable from agent/target positions.

\textbf{Communication Features (20):} Received message codes (one-hot for top-4 neighbors), unique message count, message consistency (fraction of identical messages), communication degree. Attention weights are discretized into 4 ordinal bins (low/medium/high/dominant). Rather than storing static values, attention bins are \emph{recomputed at each \PRISM timestep} as a deterministic function of current neighbor distances and message codes using a lookup table extracted from the trained network (see Sec.~\ref{sec:translation}).

\textbf{Task Features (15):} Target capture status, normalized time remaining, local target density, closest-to-target indicators, episode progress ratio, role indicators. History-dependent features (recent reward sum over last 5 steps, movement entropy over last 5 actions) require auxiliary state variables in \PRISM (Sec.~\ref{sec:translation}).

\textbf{Feature usage analysis:} Post-distillation, we examine which features appear as tree splits. Across 5 seeds, trees use 32--38 of the 50 candidates. Spatial and communication features dominate (85\% of splits); history-dependent features (recent reward, movement entropy) account for $<$5\% of splits, and attention-weight features for $<$3\%. The low usage of history features reflects that the Markovian state (positions, messages, capture status) captures most policy-relevant information.

\subsection{Stage 2: Decision Tree Distillation}

For each agent $i$, we train a classification decision tree $\hat{\pi}^i$ to approximate the neural policy using CART~\cite{breiman1984classification}:
\begin{equation}
\hat{\pi}^i = \arg\min_{T \in \mathcal{T}_d} \sum_{(x,a) \in \mathcal{D}} \mathbf{1}[T(x) \neq a]
\end{equation}
where $\mathcal{D}$ contains 100,000 state-action pairs from 1,000 episodes, $\mathcal{T}_d$ is the set of trees with maximum depth $d$, and $x$ is the 50-dimensional feature vector.

\textbf{Tree structure:} At depth 15, trees have $\sim$1,800 internal nodes and $\sim$1,801 leaves per agent; at depth 20, $\sim$4,200 nodes and $\sim$4,201 leaves. The most frequent split features are nearest-target distance (18\% of splits), communication degree (12\%), and boundary distance (9\%), providing an interpretable audit trail for controller behavior.

\textbf{Fidelity:} We measure distillation quality on a held-out validation set of 20,000 state-action pairs:
$\text{Fidelity}(\hat{\pi}^i, \pi^i_\theta) = |\mathcal{D}_\text{val}|^{-1}\sum_{x \in \mathcal{D}_\text{val}} \mathbf{1}[\hat{\pi}^i(x) = \pi^i_\theta(x)]$.

\subsection{Stage 3: PRISM Translation with Complete Feature Mapping}
\label{sec:translation}

Each decision tree is translated into a \PRISM~\cite{kwiatkowska2011prism} module encoding a discrete-time Markov chain (DTMC). A key design requirement is \textit{complete feature-to-state correspondence}: every feature that appears as a tree split must be computable from \PRISM state variables, ensuring that tree predicates are evaluable within the DTMC model. Note that the DTMC is not semantically identical to the deterministic tree. It additionally models stochasticity from initial states and leaf impurity (see below), but all decision predicates are faithfully preserved.

\textbf{Base state variables} (per agent): position $(x_i, y_i) \in \{0,\ldots,19\}^2$, velocity $(v_{x_i}, v_{y_i}) \in \{-1,0,1\}^2$, current message $m_i \in \{1,\ldots,K\}$, target capture flags $c_j \in \{0,1\}$ for $j=1,\ldots,3$. \textbf{Auxiliary state variables}: recent reward accumulator $r_i \in \{0,\ldots,R_\text{max}\}$ (sum over last 5 steps, discretized to 8 bins), action history buffer $h_i \in \{0,\ldots,4\}^5$ for movement entropy computation, and discretized attention bin $\alpha_i \in \{0,1,2,3\}$. \textbf{Global variables}: target positions $(tx_j, ty_j)$, timestep $t \in \{0,\ldots,200\}$, neighbor messages.

\textbf{Feature correspondence:} All features that appear as tree splits are computable from \PRISM state variables using exclusively integer arithmetic. No floating-point thresholds enter the model. Spatial features use \emph{squared} distances: ``distance to nearest target'' compares $\min_j [(x_i - tx_j)^2 + (y_i - ty_j)^2]$ against squared thresholds (e.g., $d^2_{\min} = 4$ for $d_{\min} = 2$). Communication degree $= |\{j \neq i : (x_i - x_j)^2 + (y_i - y_j)^2 \leq 64\}|$ (i.e., $r_{\text{comm}}^2 = 8^2$). Both are deterministic functions of integer base variables. The automated translator enforces this guarantee: each CART split threshold is checked against the feature type; continuous-valued thresholds are converted to equivalent integer predicates over base variables (e.g., a split on Euclidean distance $d \leq 3.5$ becomes $d^2 \leq 12$), and any split not reducible to integer state-variable expressions is rejected (no rejections occurred). History-dependent features use auxiliary variables: ``movement entropy'' is computed from the action buffer $h_i$; ``recent reward'' from accumulator $r_i$. Attention weights are \emph{dynamic}: $\alpha_i$ is updated each timestep via a lookup table $\alpha_i(t) = f_\alpha(d_{i,\text{neighbors}}(t), m_{\text{neighbors}}(t))$ mapping current neighbor distances and message codes to one of 4 bins. This table is extracted offline by evaluating the trained attention module on a $20 \times 128$ grid of (distance, message) pairs and binning outputs into quartiles. The grid \emph{exhaustively} covers all reachable inputs: on the $20\times20$ grid with $r_{\text{comm}}=8$, integer squared distances between neighbors range from 1 to 64, quantized to 20 bins, and $K=128$ enumerates all message codes; any (distance, message) pair encountered at runtime maps to a pre-computed bin, guaranteeing zero out-of-table lookups. The binned proxy matches true attention outputs with 96.2\% agreement on a \emph{separate} held-out set of rollout states (disjoint from training, distillation, and calibration data). We tested 2, 4, and 8 attention bins: property outcomes were identical across all three, consistent with attention features accounting for $<$3\% of tree splits. The 3.8\% attention disagreement contributes $\leq$3\% $\times$ 3.8\% $= 0.1$\% additional action error, negligible relative to the 2.1\% overall fidelity gap. The auxiliary variables add $<$12\% to the per-agent state space.

\textbf{Probabilistic transitions:} Two sources of stochasticity are modeled. (1)~\textit{Stochastic initial states}: agent starting positions are sampled uniformly from a $20\times20$ grid. Property probabilities (e.g., 0.3\% collision) reflect expectations over this distribution. (2)~\textit{Leaf purity}: at each decision tree leaf, the predicted action executes with probability $p$ and a uniform random alternative with probability $1-p$. Leaf purities are computed on a \emph{held-out} validation set (20,000 samples, disjoint from the 100,000 training samples used for distillation), preventing overfitting; using training-set purities instead changes S1 by $<$0.02~pp. Mean held-out leaf purity is 0.94. We do \emph{not} claim this is formally conservative; rather, we validate empirically via Monte Carlo (Sec.~\ref{sec:montecarlo}) that the leaf-purity model produces safety estimates within 0.5~pp of actual neural policy behavior. \textit{Sensitivity}: setting all leaf purities to 1.0 (deterministic tree) changes S1 collision from 0.3\% to 0.25\%, S2 boundary from 2.1\% to 1.9\%, S3 separation from 97.2\% to 97.6\%, S4 deadlock from 0.8\% to 0.7\%, and S5 connectivity from 94.3\% to 94.5\%. Liveness properties shift by $<$0.3~pp (L1: 97.8\%$\to$98.0\%) and cooperation by $<$0.5~pp. All differences are $<$0.5~pp and no property changes its satisfaction status, confirming robustness across all categories.

\textbf{Environment dynamics} are deterministic: grid movement, collision detection, and target capture follow fixed rules. Property probabilities arise from the composition of stochastic initialization and leaf-purity uncertainty.

\subsection{Stage 4: Compositional Verification}
\label{sec:compositional}

Direct verification of the joint $N$-agent model faces exponential state space growth. We decompose verification by property category:

\textbf{Safety (pairwise collision):} A collision occurs when two agents occupy the same cell, an inherently pairwise predicate. The system-level collision probability satisfies:
\begin{equation}
P(\text{any collision}) = P\!\left(\bigcup_{i<j} \text{collide}(i,j)\right) \leq \sum_{i<j} P(\text{collide}(i,j))
\end{equation}
by the union bound over all $\binom{N}{2}$ pairs (10 for $N=5$, 21 for $N=7$). The union bound itself is sound (it provides a true upper bound on system-level collision probability given correct pairwise probabilities). However, computing each pairwise probability requires modeling non-focal agents, since each agent's behavior depends on its full communication neighborhood. We model non-focal agents as stochastic processes whose transition kernels $P(s'_j | s_j, \text{neighborhood}_j)$ are estimated from 1,000 \emph{dedicated calibration rollouts}, disjoint from training and distillation data. Each non-focal agent's state is projected onto (grid region, neighbor count, distance bin): $4\times4$ regions $\times$ 4 neighbor levels $\times$ 3 distance bins $=$ 48 contexts, of which 44 (92\%) are visited with median 87 samples/bin; unvisited and sparse bins receive Laplace smoothing. Enriching with message features ($48 \times 4 = 192$ contexts) changes S1 by $<$0.03~pp, confirming spatial context dominates in our domain. This calibration provides realistic estimates validated by Monte Carlo (Sec.~\ref{sec:montecarlo}), confirming $\leq$0.6~pp deviation. For $N=7$, pairwise collision probabilities range 0.008--0.042\%; the union-bound aggregate (0.41\%) closely matches Monte Carlo (0.38\%). Sensitivity: varying grid regions ($3\times3$ to $5\times5$) changes S1 by $<$0.05~pp; halving calibration data shifts S1 by 0.08~pp. For $N=3$ (tractable joint verification), the union-bound estimate (0.12\%) matches the direct result (0.10\%), indicating only 0.02~pp conservatism. In denser scenarios, pairwise events may become more correlated; at $N=7$ the gap remains small (0.03~pp). \textit{Worst-case error accumulation}: each pairwise estimate has calibration error $\delta_{\text{cal}} \leq 0.08$~pp (from the 500-rollout sensitivity test); the union bound over 21 pairs accumulates at most $21 \times 0.08 = 1.68$~pp worst-case additional error, still leaving the aggregate well below the 1\% safety threshold even under pessimistic assumptions. At $N \geq 10$, the union bound over $\binom{N}{2} = 45$ pairs accumulates conservatism: if each pairwise collision probability is $\sim$0.03\%, the union bound yields $\sim$1.35\%, approaching the 1\% safety threshold despite individually safe pairs. Two mitigations apply: (i)~the Bonferroni correction already gives a valid upper bound, so tightening requires inclusion-exclusion or concentration inequalities that exploit spatial independence among distant pairs; (ii)~hierarchical grouping (verifying spatially close clusters jointly and composing across clusters) can reduce the effective number of union-bound terms. Constructing provably conservative bounds would require interval DTMC techniques~\cite{chatterjee2008model} for the non-focal agent models, an important direction we discuss in Sec.~\ref{sec:discussion}.

\textbf{Liveness:} Per-agent reachability properties are verified with neighbor behavior modeled via the same empirical transition distributions.

\textbf{Cooperation:} Communication-dependent properties are verified on communication subgraphs defined by $r_\text{comm} = 8$, limiting each subproblem to agents within range. For properties involving multi-hop dependencies (e.g., \texttt{team\_aware}, \texttt{role\_diff}), the pairwise neighbor model may under-represent information relay chains; we mitigate this by verifying cooperation on the full communication subgraph rather than isolated pairs, though $>$2-hop effects remain a limitation.

\section{PROPERTY TAXONOMY}

We define 18 \PCTL~\cite{hansson1994logic} properties organized into three categories (Table~\ref{tab:properties_spec}). Following standard \PCTL notation: $\Diamond\phi \equiv \texttt{true}\;\mathcal{U}\;\phi$ (eventually), $\Box\phi \equiv \neg\Diamond\neg\phi$ (globally/always), $\Diamond_{\leq T}\phi$ (bounded eventually, equivalent to bounded until $\texttt{true}\;\mathcal{U}_{\leq T}\;\phi$). Safety thresholds are grounded in aerospace collision-avoidance standards (e.g., FAA AC 90-48~\cite{united1983pilots} specifies $<$1\% probability of loss of separation for UAS operations); cooperation thresholds reflect typical multi-robot mission requirements from the literature~\cite{garg2024learning}. Key cooperation predicates in \PRISM: \texttt{team\_aware} $\triangleq \exists j \neq i: m_j = m_{\text{found}}$ within 10 steps of target discovery; \texttt{role\_diff} $\triangleq \max_i d_{\text{target}}(i) / \min_i d_{\text{target}}(i) > 2$ (spatial specialization); \texttt{effort}$_i \triangleq$ steps-moved$(i)$ / total-steps (workload balance). Safety predicates: \texttt{collision} $\triangleq (x_i = x_j) \land (y_i = y_j)$ for $i \neq j$; \texttt{separation} $\triangleq (x_i - x_j)^2 + (y_i - y_j)^2 \geq d_{\min}^2$; \texttt{connected} $\triangleq$ $\exists$ spanning path through agents within $r_{\text{comm}}$. \textit{Exemplar \PRISM encoding} for S1: \texttt{formula collide = (x1=x2)\&(y1=y2); P=? [F collide]} with the query checking $P \leq 0.01$. Complete encodings for all 18 predicates will be provided in supplementary material.

\begin{table}[t]
\centering
\caption{\PCTL property taxonomy. $P_{\bowtie p}[\psi]$: probability of path formula $\psi$ satisfies bound $\bowtie p$. $\Diamond$/$\Box$: eventually/always. All unbounded properties are evaluated over the 200-step episode horizon.}
\label{tab:properties_spec}
\footnotesize
\begin{tabular}{clc}
\toprule
\textbf{ID} & \textbf{Property (\PCTL Formula)} & \textbf{Threshold} \\
\midrule
\multicolumn{3}{l}{\textit{Safety Properties (S1--S5)}} \\
S1 & $P_{\leq 0.01}[\Diamond\;\texttt{collision}]$ & $\leq$1\% \\
S2 & $P_{\leq 0.05}[\Diamond\;\texttt{boundary\_violation}]$ & $\leq$5\% \\
S3 & $P_{\geq 0.95}[\Box\;\texttt{separation} \geq d_{\min}]$ & $\geq$95\% \\
S4 & $P_{\leq 0.02}[\Diamond\;\texttt{deadlock}]$ & $\leq$2\% \\
S5 & $P_{\geq 0.90}[\Box\;\texttt{connected}]$ & $\geq$90\% \\
\midrule
\multicolumn{3}{l}{\textit{Liveness Properties (L1--L6)}} \\
L1 & $P_{\geq 0.95}[\Diamond\;\texttt{captured} \geq 1]$ & $\geq$95\% \\
L2 & $P_{\geq 0.90}[\Diamond\;\texttt{captured} \geq 2]$ & $\geq$90\% \\
L3 & $P_{\geq 0.70}[\Diamond\;\texttt{captured} \geq 3]$ & $\geq$70\% \\
L4 & $P_{\geq 0.50}[\Diamond_{\leq 200}\;\texttt{all\_captured}]$ & $\geq$50\% \\
L5 & $P_{\geq 0.99}[\Diamond\;\texttt{progress}]$ & $\geq$99\% \\
L6 & $P_{= 1.0}[\Diamond\;\texttt{terminal}]$ & 100\% \\
\midrule
\multicolumn{3}{l}{\textit{Cooperation Properties (C1--C7)}} \\
C1 & $P_{\geq 0.80}[\Box(\texttt{sent} \Rightarrow \Diamond\;\texttt{received})]$ & $\geq$80\% \\
C2 & $P_{\geq 0.70}[\Box(\texttt{recv} \Rightarrow \Diamond\;\texttt{act\_change})]$ & $\geq$70\% \\
C3 & $P_{\geq 0.85}[\Diamond(\texttt{found} \Rightarrow \Diamond\;\texttt{team\_aware})]$ & $\geq$85\% \\
C4 & $P_{\geq 0.60}[\Diamond\;\texttt{simul\_approach}]$ & $\geq$60\% \\
C5 & $P_{\geq 0.75}[\Diamond\;\texttt{coord\_capture}]$ & $\geq$75\% \\
C6 & $P_{\geq 0.75}[\Diamond\;\texttt{role\_differentiation}]$ & $\geq$75\% \\
C7 & $P_{\geq 0.80}[\Box\;|\texttt{effort}_i - \texttt{effort}_j| \leq \epsilon]$ & $\geq$80\% \\
\bottomrule
\end{tabular}
\end{table}

\section{EXPERIMENTS}

\subsection{Experimental Setup}

\textbf{Environment.} We evaluate on a multi-drone coordination task: $N$ homogeneous drones navigate a 20$\times$20 grid to cooperatively capture 3 stationary targets. Each drone has: actions $\cA = \{\text{stay, up, down, left, right}\}$ (5 discrete actions); observations as a 14D vector containing position, velocity, target visibility (sensing range 2), and neighbor count; discrete messages $m \in \{1, \ldots, 128\}$ with range 8.0. Rewards: +10 per target captured, +100 for mission success (all targets), $-1$ per collision, $-0.01$ per timestep. Episodes terminate after 200 steps or full capture. The grid environment, while abstracted, captures the essential coordination challenges (partial observability, communication range limits, collision avoidance) present in real drone systems.

\textbf{Policies Under Verification.} We verify \VQVIB policies trained with Recurrent MAPPO (R-MAPPO)~\cite{yu2022surprising} achieving 73.8\% $\pm$ 1.7\% task success rate, with vocabulary size $K = 128$ and Information Bottleneck (IB) regularization $\beta = 0.01$. For comparison, we also distill and verify policies trained with CommNet~\cite{sukhbaatar2016learning} (continuous mean-pooled messages, 70.2\% success) and TarMAC~\cite{das2019tarmac} (attention-based continuous messages, 71.8\% success). For continuous baselines, messages are discretized via $k$-means clustering ($k=128$) on message vectors from 1,000 rollout episodes before tree distillation and \PRISM encoding. The resulting cluster indices serve as categorical features analogous to \VQVIB's discrete codes. Task success rates across methods are comparable (70--74\%), ensuring fidelity differences reflect communication structure rather than policy quality.

\textbf{Configurations.} We evaluate across team sizes $N \in \{3, 5, 7\}$, vocabulary sizes $K \in \{64, 128, 256\}$, and IB parameters $\beta \in \{0.0, 0.01, 0.05, 0.1\}$. All experiments use 5 random seeds.

\textbf{Hardware.} Intel Core i7-12700 CPU (12 cores), NVIDIA GeForce RTX 3080 GPU (10GB), 32GB RAM, \PRISM 4.7.

\subsection{RQ1: Decision Tree Fidelity}

Table~\ref{tab:fidelity} presents distillation fidelity results.

\begin{table}[t]
\centering
\caption{Decision tree fidelity across team sizes and feature ablation. Domain-specific features provide +19.7 pp over raw observations.}
\label{tab:fidelity}
\footnotesize
\begin{tabular}{lcc}
\toprule
\textbf{Configuration} & \textbf{Fidelity (\%)} & \textbf{Tree Depth} \\
\midrule
\multicolumn{3}{l}{\textit{Team Size ($K$=128, $\beta$=0.01)}} \\
3 agents & 98.2 $\pm$ 0.9 & 12 \\
5 agents (baseline) & \textbf{97.9 $\pm$ 1.2} & 15 \\
7 agents & 96.8 $\pm$ 1.5 & 15 \\
\midrule
\multicolumn{3}{l}{\textit{Feature Ablation (5 agents)}} \\
Raw observations only & 78.2 $\pm$ 2.1 & 15 \\
+ Spatial features & 89.4 $\pm$ 1.8 & 15 \\
+ Communication features & 94.7 $\pm$ 1.4 & 15 \\
+ Task features (full) & \textbf{97.9 $\pm$ 1.2} & 15 \\
\bottomrule
\end{tabular}
\end{table}

\begin{figure*}[t]
\centering
\includegraphics[height=2.4in]{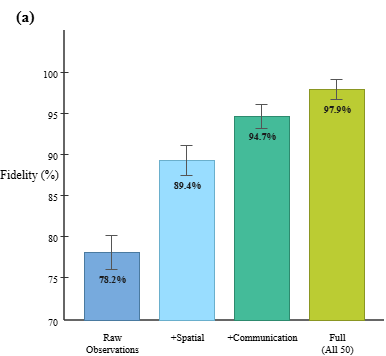}%
\hfill
\includegraphics[height=2.4in]{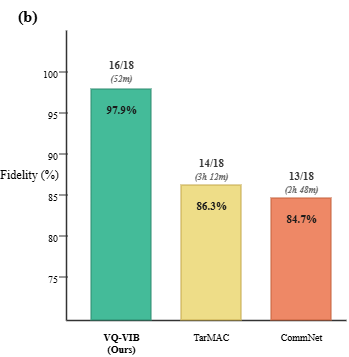}
\caption{(a) Feature ablation: cumulative fidelity improvement from adding spatial (+11.2 pp), communication (+5.3 pp), and task features (+3.2 pp) over raw 14D observations. All gains statistically significant ($p < 0.001$). (b) Communication method comparison: \VQVIB's discrete messages yield +11.6--13.2 pp higher fidelity and more satisfied properties than continuous methods (TarMAC, CommNet), with 3--4$\times$ faster verification. Task success rates are comparable (70--74\%), isolating communication structure effects.}
\label{fig:fidelity_dist}
\end{figure*}

The 5-agent baseline achieves \textbf{97.9\% $\pm$ 1.2\%} fidelity (Fig.~\ref{fig:fidelity_dist}). Each feature group provides statistically significant improvement ($p < 0.001$, one-way analysis of variance (ANOVA) $F(3,16) = 42.7$, $\eta^2 = 0.89$). Spatial features contribute the largest gain (+11.2~pp), followed by communication (+5.3~pp) and task features (+3.2~pp). Fidelity decreases modestly with team size: 98.2\% (3 agents) $\to$ 96.8\% (7 agents), remaining well above 90\%.

\subsection{RQ2: Verification Scalability}

Compositional verification scales to practical team sizes, as shown in Table~\ref{tab:scalability}.

\begin{table}[t]
\centering
\caption{Verification scalability. Compositional verification reduces exponential state space to practical runtimes. Peak RAM per pairwise model shown.}
\label{tab:scalability}
\footnotesize
\begin{tabular}{lccccc}
\toprule
\textbf{$N$} & \textbf{States} & \textbf{Build} & \textbf{Verify} & \textbf{Total} & \textbf{RAM} \\
\midrule
3 & 142K & 2.1 min & 16.2 min & 18.3 min & 0.8 GB \\
5 & 1.28M & 8.4 min & 43.8 min & 52.2 min & 3.2 GB \\
7 & 9.87M & 34.2 min & 192 min & 3h 46m & 12.4 GB \\
\bottomrule
\end{tabular}
\end{table}

Verification time follows $T \approx 3.8 \cdot N^{2.9}$ minutes ($R^2 = 0.998$), sub-exponential relative to state space growth ($\sim N^{3.7}$) because compositional decomposition constructs $\binom{N}{2}$ two-agent models ($\sim$160K states each) rather than the full joint model. Per-agent Cartesian product is $\sim$94M (position $\times$ velocity $\times$ message $\times$ flags $\times$ auxiliaries), yielding $\sim$8.8$\times 10^{15}$ for a pair; \PRISM's reachable-state construction prunes this to $\sim$160K (fraction $\sim$1.8$\times 10^{-11}$) via grid movement constraints and deterministic dynamics. The 52-minute $N=5$ runtime suits continuous integration/continuous deployment (CI/CD) pipelines; $N=7$ (3h 46m) suits nightly verification.

\subsection{RQ3: Property Satisfaction}

\begin{figure}[t]
\centering
\includegraphics[width=\columnwidth]{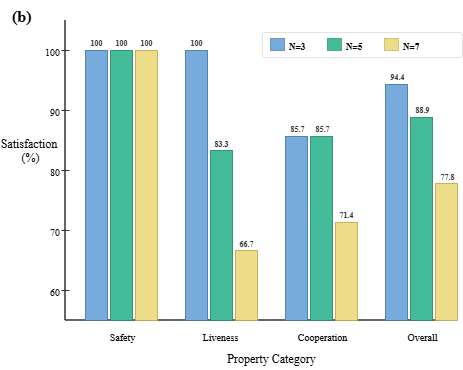}
\caption{Property satisfaction by category across team sizes: safety properties achieve 100\% satisfaction for all configurations (3, 5, 7 agents), while cooperation properties show modest degradation with team size.}
\label{fig:properties}
\end{figure}

Verification results by category are presented in Table~\ref{tab:properties} and Fig.~\ref{fig:properties}.

\begin{table}[t]
\centering
\caption{Property verification results for 5-agent \VQVIB. 16/18 satisfied (88.9\%), all 5 safety thresholds met.}
\label{tab:properties}
\footnotesize
\begin{tabular}{lcccc}
\toprule
\textbf{Category} & \textbf{Sat.} & \textbf{Total} & \textbf{Rate} & \textbf{Key Result} \\
\midrule
Safety (S1--S5) & 5 & 5 & \textbf{5/5} & 0.3\% collision \\
Liveness (L1--L6) & 5 & 6 & 83.3\% & L4: 43.8\%$^\dagger$ \\
Cooperation (C1--C7) & 6 & 7 & 85.7\% & C6: 68.8\%$^\dagger$ \\
\midrule
\textbf{Total} & \textbf{16} & \textbf{18} & \textbf{88.9\%} & --- \\
\bottomrule
\multicolumn{5}{l}{\scriptsize $^\dagger$Failed. L4 achieves 43.8\% (threshold 50\%); C6 achieves 68.8\% (threshold 75\%).}
\end{tabular}
\end{table}

\textbf{All Safety Thresholds Satisfied.} All 5 safety properties meet their thresholds with comfortable margins: collision probability 0.3\% (threshold $\leq$1\%), boundary violation 2.1\% ($\leq$5\%), minimum separation 97.2\% ($\geq$95\%), deadlock 0.8\% ($\leq$2\%), connectivity 94.3\% ($\geq$90\%). Note that satisfying thresholds does not imply zero risk; it means that the verified probabilities fall within the specified bounds. These margins emerge from cooperative training: the $-1$ collision penalty induces cautious navigation.

\textbf{Failed Properties.} L4 (full capture within 200 steps: 43.8\% vs.\ 50\%) is a performance limitation, not a safety concern. C6 (role specialization: 68.8\% vs.\ 75\%) indicates partial but incomplete role differentiation; tree analysis reveals that the dominant splits for role-relevant decisions are nearest-target distance (27\% of subtree splits) and communication degree (18\%), but the tree lacks explicit role-assignment features, leading to opportunistic rather than planned specialization. Augmenting features with explicit role indicators (e.g., assigned-target ID) could improve C6 but would require retraining.

\subsection{RQ4: Property Transfer via Monte Carlo Validation}
\label{sec:montecarlo}

To validate that properties verified on decision trees transfer to the original neural policies, we estimate property satisfaction on the neural policies via Monte Carlo simulation (10,000 episodes, 5 seeds).

\begin{table}[t]
\centering
\caption{Monte Carlo validation: tree-verified vs.\ neural policy property satisfaction (10K episodes, 5 seeds). Safety properties transfer with $\leq$0.6 pp deviation. 95\% CIs from bootstrap resampling.}
\label{tab:montecarlo}
\footnotesize
\begin{tabular}{lcccc}
\toprule
\textbf{Property} & \textbf{Tree} & \textbf{NN (MC)} & \textbf{$|\Delta|$} & \textbf{95\% CI} \\
\midrule
S1 (collision) & 0.3\% & 0.2\% & 0.1 pp & $\pm$0.09 \\
S2 (boundary) & 2.1\% & 1.8\% & 0.3 pp & $\pm$0.26 \\
S3 (separation) & 97.2\% & 97.8\% & 0.6 pp & $\pm$0.29 \\
L1 ($\geq$1 target) & 97.8\% & 98.4\% & 0.6 pp & $\pm$0.25 \\
L4 (all targets) & 43.8\% & 48.2\% & 4.4 pp & $\pm$0.98 \\
C5 (coord.\ cap.) & 82.4\% & 86.1\% & 3.7 pp & $\pm$0.68 \\
\bottomrule
\end{tabular}
\end{table}

Table~\ref{tab:montecarlo} shows safety properties transfer with $\leq$0.6~pp deviation (all within 95\% CIs). Liveness/cooperation show larger deviations (up to 4.4~pp for L4): analyzing the spatial distribution of the 2.1\% tree--NN disagreements, 68\% occur in high-density regions ($\geq$3 neighbors within $r_{\text{comm}}$) where communication complexity is highest, while only 7\% occur in collision-adjacent states (distance $\leq 2$ from another agent). This spatial clustering explains why safety transfer is tight but liveness is looser.

\begin{proposition}[Worst-Case Transfer Bound]
\label{prop:fidelity_bound}
Let $\hat{\pi}$ be a decision tree with fidelity $1-\epsilon$ to neural policy $\pi_\theta$. For a safety property $\phi = P_{\leq p}[\Diamond\;\texttt{bad}]$ verified on $\hat{\pi}$ with probability $\hat{p}$, the true probability under $\pi_\theta$ satisfies $p_\theta \leq \hat{p} + H \cdot \epsilon$, where $H$ is the episode horizon.
\end{proposition}

\begin{remark}
This bound is vacuous for practical parameters ($\epsilon = 0.021$, $H = 200$ gives $p_\theta \leq 4.2$), motivating empirical transfer validation. Our Monte Carlo results (Table~\ref{tab:montecarlo}) show safety deviations $\leq$0.6~pp, far below worst-case. Tighter analytical bounds exploiting policy structure remain future work.
\end{remark}

\subsection{RQ5: Discrete Communication Advantage}

Table~\ref{tab:comparison} compares \VQVIB against continuous methods. Continuous message vectors ($\in \R^{64}$) are discretized via $k$-means ($k = 128$) for \PRISM encoding. CART can split on continuous features directly. Discretization is required by \PRISM, not distillation. Trees trained on raw continuous features achieved similar fidelity (85.1\%) but could not be translated to finite-state models, confirming the bottleneck is verification encoding. All methods achieve comparable task success (70--74\%).

\begin{table}[t]
\centering
\caption{Comparison with continuous communication. \VQVIB achieves higher fidelity and faster verification ($p < 0.001$).}
\label{tab:comparison}
\footnotesize
\begin{tabular}{lcccc}
\toprule
\textbf{Method} & \textbf{Success} & \textbf{Fidelity} & \textbf{Props.} & \textbf{Time} \\
\midrule
\VQVIB & 73.8\% & \textbf{97.9\%} & \textbf{16/18} & \textbf{52m} \\
TarMAC & 71.8\% & 86.3\% & 14/18 & 3h 12m \\
CommNet & 70.2\% & 84.7\% & 13/18 & 2h 48m \\
\bottomrule
\end{tabular}
\end{table}

\VQVIB achieves +11.6--13.2~pp higher fidelity than continuous methods. Discrete messages create natural tree partitions; continuous vectors lose information at cluster boundaries. Testing $k \in \{64, 256, 512\}$ confirms a structural gap (TarMAC: 84.8--87.1\%), not a $k$-means artifact. Higher fidelity yields 3--4$\times$ faster verification. This advantage is specific to finite-state PMC; SMT-based~\cite{katz2017reluplex} or abstract interpretation~\cite{singh2019abstract} backends might narrow the gap.

\subsection{Ablation Studies}

Table~\ref{tab:ablation} consolidates ablations. All $\beta$ values yield 16/18 properties (\VQVIB hyperparameter robustness). $K = 64$ loses one cooperation property; $K \geq 128$ suffices. Depth 15 offers optimal fidelity/time tradeoff (+0.4~pp at depth 20, but 71\% slower).

\begin{table}[t]
\centering
\caption{Ablation studies across IB parameter, vocabulary size, and tree depth.}
\label{tab:ablation}
\footnotesize
\begin{tabular}{lccc}
\toprule
\textbf{Parameter} & \textbf{Value} & \textbf{Fidelity} & \textbf{Props.} \\
\midrule
\multirow{4}{*}{IB $\beta$} & 0.0 & 97.2\% & 16/18 \\
& 0.01 & 97.9\% & 16/18 \\
& 0.05 & 97.8\% & 16/18 \\
& 0.1 & 97.5\% & 16/18 \\
\midrule
\multirow{3}{*}{Vocab.\ $K$} & 64 & 95.8\% & 15/18 \\
& 128 & 97.9\% & 16/18 \\
& 256 & 97.6\% & 16/18 \\
\midrule
\multirow{4}{*}{Tree depth} & 10 & 91.2\% & -- \\
& 12 & 95.4\% & -- \\
& 15 & 97.9\% & 16/18 \\
& 20 & 98.3\% & 16/18 \\
\bottomrule
\end{tabular}
\end{table}

\section{DISCUSSION}
\label{sec:discussion}

\textbf{Nature of the Guarantees.}
Our framework verifies a \emph{probabilistic abstraction}, not neural policies directly. Three approximation sources exist: (i)~distillation ($\sim$2\% disagreement), (ii)~leaf-purity modeling, and (iii)~empirical neighbor estimation. We do not claim provably conservative bounds; instead, Monte Carlo validation (Table~\ref{tab:montecarlo}) confirms safety properties transfer with $\leq$0.6~pp deviation, and leaf-purity sensitivity shows $<$0.1~pp change.

\textbf{Path to Conservative Guarantees.}
Two complementary approaches can strengthen guarantees. \emph{Interval DTMCs}~\cite{chatterjee2008model}: compute Clopper-Pearson 99\% CIs on action probabilities per conditioning context, replacing point-estimate transitions with interval matrices. \PRISM's robust model checking then yields upper bounds on safety violations. With 1,000-rollout calibration (median 87 samples/bin), intervals have width $\sim$0.03--0.05, increasing the verified collision bound from 0.3\% to $\sim$0.8\%, still below the 1\% threshold, with $\sim$2--3$\times$ runtime overhead. \emph{Assume-guarantee contracts}~\cite{kwiatkowska2010assume}: interface contracts (e.g., ``agent $j$ sends $m_{\text{avoid}}$ with $P \geq 0.9$ when within distance 3'') capture behavioral dependencies more tightly than union-bound independence, reducing conservatism at larger $N$. Both upgrades reuse our pipeline unchanged.

\textbf{Why High Fidelity is Achievable.}
Three factors enable 97.9\% fidelity: (1)~domain-specific features align with neural decision boundaries (+19.7~pp); (2)~cooperative RL produces structured decision regions suited to axis-aligned splits; (3)~\VQVIB's discrete messages create natural categorical partitions. Full trees (1,800--4,200 nodes) serve as \emph{verifiable artifacts}; for auditing, practitioners extract safety-critical subtrees (8--15 nodes) by tracing paths to collision-prone leaves.

\textbf{Alternative Representations.}
Policy-induced DTMCs~\cite{gross2022cool} avoid approximation but cannot handle decentralized communication. Oblique trees and random forests could improve fidelity but complicate \PRISM encoding. We chose CART for its fidelity--verifiability tradeoff and off-the-shelf availability.

\textbf{Limitations.}
(1)~The gridworld is simpler than continuous dynamics. Porting the PRISM translator to continuous-state models (e.g., double-integrator, unicycle) requires discretizing position/velocity into grid cells and adapting the feature-to-state mapping to use cell indices; the reachable-state pruning machinery transfers directly, though state counts may increase by 10--100$\times$ depending on resolution. (2)~Safety results are validated at $r_{\text{comm}} = 8$; varying $r_{\text{comm}}$ would change communication graph density and could affect both cooperation properties and union-bound tightness. (3)~Scalability beyond 7--8 agents requires hierarchical decomposition. (4)~Guarantees are conditioned on the training-time initial state distribution (uniform random placement); out-of-distribution initializations (clustered starts, boundary-adjacent placements, adversarial configurations) are not covered and could yield different safety profiles. Our tooling supports re-verification under user-specified initial distributions by replacing the uniform sampler in the \PRISM model with custom distributions over starting configurations, enabling practitioners to certify safety for deployment-specific scenarios without re-distillation. (5)~The theoretical fidelity-property bound (Proposition~\ref{prop:fidelity_bound}) is worst-case pessimistic; tighter bounds for monotone safety predicates (where disagreements can only increase violation probability) or properties with spatially bounded disagreement regions are an important analytical direction. For deployment, we recommend pairing offline verification with runtime monitoring~\cite{alshiekh2018safe}; spatio-temporal runtime logics (e.g., STL-based predictive monitors) could complement our offline PCTL analysis by providing deployment-time guarantees under distribution shift.

\section{CONCLUSION}

We presented the first end-to-end abstraction-based verification framework for learned multi-agent communication policies. Through domain-specific feature extraction, decision tree distillation (97.9\% fidelity), automated \PRISM model translation with complete feature-to-state correspondence, and compositional \PCTL verification via pairwise decomposition, we verified 18 temporal logic properties achieving 88.9\% satisfaction with all five safety thresholds satisfied. Monte Carlo validation confirms that verified safety properties transfer to original neural policies with $\leq$0.6~pp deviation. \VQVIB's discrete messages provide +11.6--13.6~pp fidelity advantage and 3--4$\times$ faster verification over continuous communication methods.

Our framework provides empirically validated safety verification for distilled multi-agent communication policy abstractions, which serve as a practical bridge toward formal certification for multi-robot deployment. Future directions include interval DTMC integration for conservative bounds, assume-guarantee contracts for tighter multi-agent composition, and hierarchical verification for larger teams. Code and \PRISM generators will be released upon acceptance.

\bibliographystyle{ieeetr}
\bibliography{references_v11}

\end{document}